\documentclass{article}
\usepackage{arxiv}
\usepackage{times}
\usepackage{soul}
\usepackage{url}
\usepackage[hidelinks]{hyperref}
\usepackage[utf8]{inputenc}
\usepackage[small]{caption}
\usepackage{graphicx}
\usepackage{amsmath}
\usepackage{amssymb}
\usepackage{amsthm}
\usepackage{booktabs}
\usepackage{algorithm}
\usepackage{algorithmic}
\usepackage{subcaption}
\usepackage{makecell}
\usepackage{enumitem}
\usepackage[table]{xcolor}
\definecolor{groupgray}{gray}{0.96}
\definecolor{ourblue}{RGB}{240,250,255}

\title{ESPO: Entropy Importance Sampling Policy Optimization}

\author{
Yuepeng Sheng$^1$$^*$
\and
Yuwei Huang$^1$\thanks{Equal contribution.}\and
Shuman Liu$^1$\and
Anxiang Zeng$^1$\And
Haibo Zhang$^1$\thanks{Corresponding author.}\\
\affiliations
$^1$LLM Team, Shopee Pte. Ltd.\\
\emails
\{yuepeng.sheng, yuwei.huang, liushuman, peter.wu\}@shopee.com,
zeng0118@ntu.edu.sg
}

\begin{document}

\maketitle

\newcommand{\ourpo}{ESPO}
\begin{abstract}
    Reinforcement learning (RL) has become a central component of post-training for large language models (LLMs), particularly for complex reasoning tasks that require stable optimization over long generation horizons.
    However, achieving performance at scale often introduces a fundamental trade-off between training stability and training efficiency. Token-level optimization applies fine-grained updates at the individual units, but is prone to high variance in gradient estimation, which can result in unstable training dynamics.
    In contrast, Sequence-level optimization often relies on aggressive clipping mechanisms to ensure stable updates.
    However, such design may discard a large fraction of valid training samples, leading to inefficient gradient utilization and reduced training efficiency.
    We refer to this phenomenon as \emph{gradient underutilization}.
    In this work, we propose \textbf{E}ntropy Importance \textbf{S}ampling \textbf{P}olicy \textbf{O}ptimization (\ourpo), a novel framework that aims to combine fine-grained updates with stable training. \ourpo~decomposes sequences into groups based on predictive entropy, enabling (1) Entropy Grouping Importance Sampling to capture intra-sequence heterogeneity, and (2) Entropy Adaptive Clipping to dynamically allocate trust regions based on model uncertainty. Extensive experiments on mathematical reasoning benchmarks demonstrate that \ourpo~not only accelerates convergence but also achieves state-of-the-art performance, notably improving accuracy on the challenging mathematical benchmarks.\footnote{Code is available at: \url{https://github.com/ShopeeLLM/ESPO}}
\end{abstract}

\section{Introduction}

\begin{figure}
    \centering
    \includegraphics[width=0.5\textwidth]{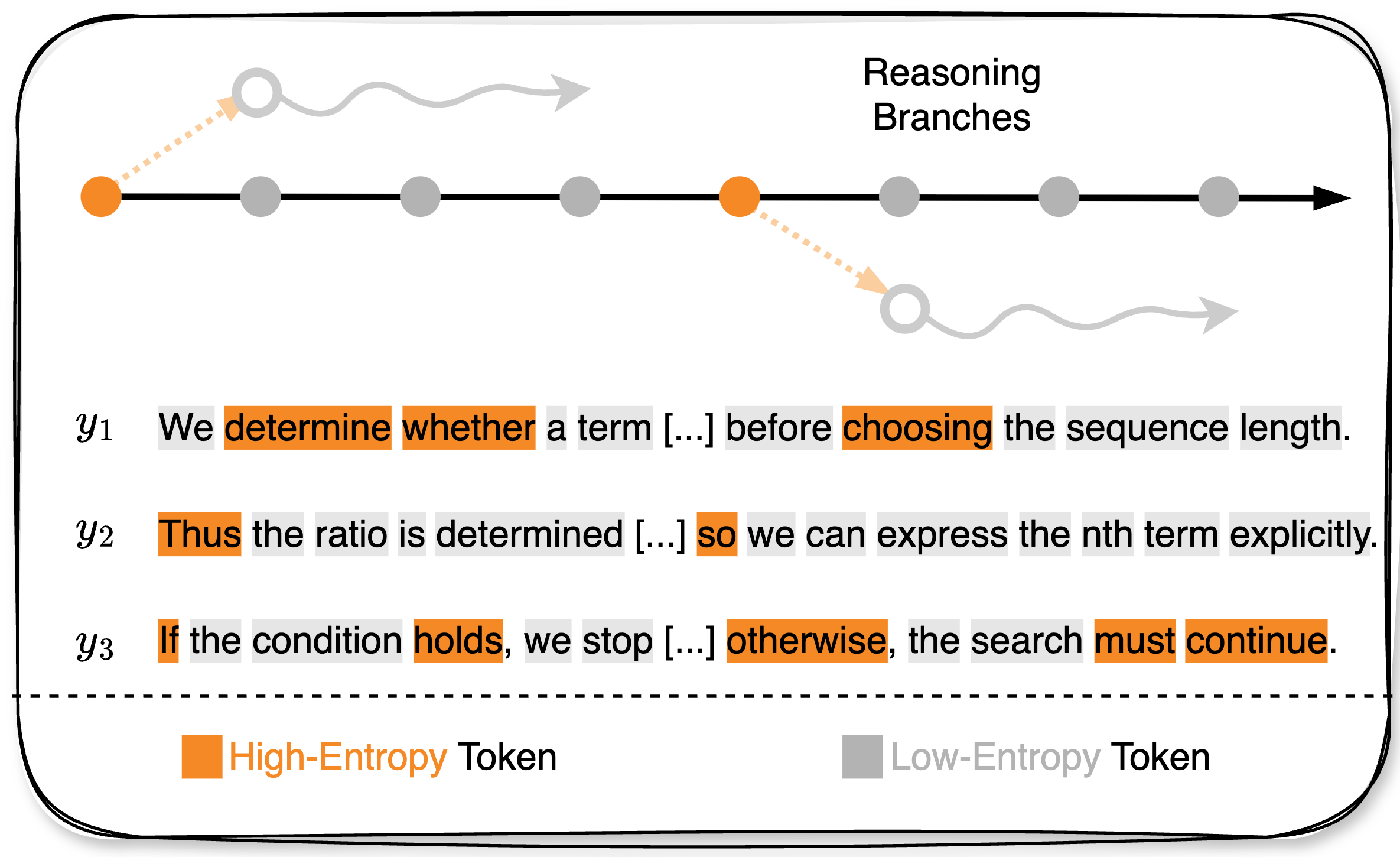}
    \caption{Illustration of high-entropy tokens as decision points that guide reasoning toward different trajectories.}
    \label{fig:ent_branches}
\end{figure}

Recent progress in LLM has increasingly emphasized their ability to perform complex reasoning, where correct outputs depend on multi-step logical inference rather than surface-level pattern matching. In such settings, post-training objectives extend beyond instruction following and require aligning the model’s generation process with verifiable reasoning correctness. Reinforcement learning has therefore become a key component of modern reasoning-oriented training pipelines.
By allowing models to explore diverse reasoning trajectories and optimize towards task-level correctness, RL provides a natural framework for improving decision-making over long sequences.

However, current RL strategies face a fundamental trade-off induced by \textit{optimization granularity}, which directly affects the balance between training stability and training efficiency.
 Early approaches like Proximal Policy Optimization (PPO) \cite{schulman2017proximal} operate at the token level, offering fine-grained control but suffering from high variance and the heavy computational burden of maintaining value networks. To address this, a series of variants have been proposed and can be broadly categorized into two classes: token-level (e.g., GRPO~\cite{shao2024deepseekmath}, DAPO~\cite{yu2025dapo}, GMPO~\cite{zhao2025geometric}, CISPO~\cite{chen2025minimax}, etc.) and sequence-level (e.g., GSPO~\cite{zheng2025group}, etc.) optimization. In practice, token-level variants are prone to unstable training dynamics, while sequence-level approaches tend to exhibit low data utilization due to aggressive clipping.

To resolve these limitations, we introduce \textbf{E}ntropy Importance \textbf{S}ampling \textbf{P}olicy \textbf{O}ptimization (\ourpo), a framework that restores fine-grained optimization without sacrificing the stability benefits of group-based methods.
The core insight of \ourpo~is that \textit{policy entropy~\cite{shannon1948mathematical} serves as a natural indicator for optimization granularity}. Previous work \cite{wang2025beyond} proposes using entropy as the criterion for credit assignment by computing gradients only for the most uncertain tokens. This selective update improves efficiency by concentrating the learning on regions where the model exhibits greater uncertainty. Another line of work \cite{wang2025stabilizing} categorizes tokens into distinct entropy types, and assigns each a fixed clipping range. These designs are motivated by the role of entropy, as illustrated in Figure~\ref{fig:ent_branches}, high-entropy tokens correspond to pivotal decision points that steer the reasoning trajectory across alternative branches.

\ourpo~operates on \textbf{token groups}—dynamically grouped by their entropy profile.
Specifically, it employs \textbf{Entropy Grouping Importance Sampling} to decompose the global sequence ratio into local group ratios, allowing gradients to flow freely in consistent segments while being carefully modulated in divergent ones.
Furthermore, it introduces \textbf{Entropy Adaptive Clipping}, which assigns wider clipping bounds to high-entropy (uncertain) groups to encourage exploration, and tighter bounds to low-entropy (confident) groups to enforce stability.

The contributions of this work are summarized as follows:

\begin{itemize}[leftmargin=1em,noitemsep,topsep=2pt]
    \item We identify limitations in existing methods with respect to training efficiency and training stability, and categorize these issues into token-level and sequence-level policy optimization.

    \item We propose \textbf{E}ntropy Importance \textbf{S}ampling \textbf{P}olicy \textbf{O}ptimization (\ourpo), an entropy-based algorithm that control optimization granularity and improve data utilization.

    \item We provide detailed empirical analyses and ablation studies to examine the effects of core components, offering competitive performance on mathematical reasoning benchmarks.
\end{itemize}

\section{Background}

\subsection{Related Works}

Reinforcement learning has emerged as a central paradigm for optimizing LLMs, particularly in tasks requiring verifiable reasoning such as mathematics, coding, and question answering.
Early RL approaches were largely based on Proximal Policy Optimization (PPO)~\cite{schulman2017proximal}, where policy updates are guided by reward models or rule-based verifiers.
To reduce the computational cost and instability of value-function training in PPO, Group Relative Policy Optimization (GRPO)~\cite{shao2024deepseekmathpushinglimitsmathematical} eliminates the critic and estimates advantages through group-wise relative ranking, achieving strong performance without relying on explicit value models.
Building on GRPO, a series of recent methods have further improved stability, credit assignment, and rollout efficiency.
DAPO~\cite{yu2025dapo} introduces clip-higher and dynamic sampling to counter entropy collapse and filter zero-advantage groups, while GSPO~\cite{zheng2025group} elevates optimization granularity to the sequence level, aligning the update unit with reward signals.
The geometric average of importance sampling ratios is used in GMPO~\cite{zhao2025geometric} and employs a narrower clip range to reduce the optimization variance. GTPO~\cite{tan2025gtpo} and GRPO-S leverages entropy-weighted rewards to better handle outliers and sparse credit.
DCPO~\cite{yang2025dcpo} dynamically adapts clipping ranges by token to enhance exploration and make better use of otherwise degenerate rollouts.
CISPO~\cite{chen2025minimax} further refines clipping behavior by applying clipping only to the importance sampling ratios, while leaving gradient magnitudes unclipped.
Complementary efforts focus on reward shaping, advantage normalization, and data-centric strategies: EMPO~\cite{zhang2025right} incorporates semantic entropy, BNPO~\cite{xiao2025bnpo} normalizes rewards via Beta distributions, and large-scale datasets with curriculum learning, such as Open-Reasoner-Zero~\cite{hu2025open}, have proven crucial for high-quality signal alignment. Beyond this, SPO~\cite{guo2025segment} incorporates segment-level optimization as a principled design to address the credit assignment challenge, thereby achieving more accurate reward attribution.
Together, these developments reflect a broader trend toward efficient, stable, and reasoning-aligned RL methods for LLMs.

\subsection{Preliminary}
\textbf{Token-level policy optimization} performs policy updates at the granularity of individual tokens.
Representative approaches include GRPO, DAPO, GMPO and CISPO, which differ in their data utilization method, importance sampling computation, clipping strategies. We introduce GRPO as a concrete example in the following.

GRPO~\cite{shao2024deepseekmathpushinglimitsmathematical} is a reinforcement learning algorithm that evaluates candidate responses purely through relative preferences within a group.
For a given query \(q\), GRPO generates a set of responses \(\{y_i\}_{i=1}^{G}\) and assigns a sequence-level advantage \(A_i\) to each response.
The policy is updated using a PPO-inspired objective applied at the token level:
\begin{multline}
\mathcal{J}_{\text{GRPO}}(\theta)=
\mathbb E_{x \sim \mathcal{D}, \, \{y_i\}_{i=1}^G \sim \pi_{\theta_{\text{old}}}(\cdot|x)}\Bigg[ 
\frac{1}{G}\sum_{i=1}^{G}\frac{1}{|y_i|}\sum_{t=1}^{|y_i|}
\\\min\Big(r_{i,t}(\theta)A_i,\
\operatorname{clip}(r_{i,t}(\theta),1-\epsilon_{\text{low}}^{\text{token}},1+\epsilon_{\text{high}}^{\text{token}})A_i \Big)\Bigg]
\end{multline}
where \(r_{i,t}(\theta)=\frac{\pi_{\theta}(y_{i,t}\mid q,y_{i,<t})}{\pi_{\theta_{\text{old}}}(y_{i,t}\mid q,y_{i,<t})}\) represents the token-level importance ratio.

Despite its simplicity, GRPO faces significant challenges.
Token-level ratios are inherently noisy, and this noise compounds across long sequences, leading to unstable gradients~\cite{zhao2025geometric}.
Clipping can exacerbate this instability by unevenly scaling token contributions, sometimes resulting in divergence or training collapse.
Furthermore, since each token ratio is estimated from a single sample, it provides only a weak approximation of the underlying policy distribution, misaligning the optimization with actual reward signals~\cite{zheng2025group}.
Lastly, token-focused credit assignment ignores global sequence patterns, making it difficult to guide coherent long-horizon generation.

\textbf{Sequence-level policy optimization} aims to better align the optimization objective with sequence-level importance sampling by treating entire trajectories as optimization units.
GSPO~\cite{zheng2025group} is a representative approach in this category.

Instead of treating tokens independently, GSPO computes an importance ratio at the sequence level:
\begin{equation}
s_i(\theta)=\Big(\frac{\pi_\theta(y_i\mid q)}{\pi_{\theta_{\text{old}}}(y_i\mid q)}\Big)^{\frac{1}{|y_i|}}
\end{equation}
and projects it to individual tokens using a stop-gradient operation:
\begin{equation}
s_{i,t}(\theta)=\text{sg}[s_i(\theta)]\cdot \frac{\pi_\theta(y_{i,t}\mid q,y_{i,<t})}{\text{sg}[\pi_\theta(y_{i,t}\mid q,y_{i,<t})]}
\end{equation}

The training objective combines sequence-level ratios with clipped advantages:
\begin{multline}
\mathcal{J}_{\text{GSPO}}(\theta)
=
\mathbb E_{x \sim \mathcal{D}, \, \{y_i\}_{i=1}^G \sim \pi_{\theta_{\text{old}}}(\cdot|x)}
\Bigg[
\frac{1}{G}\sum_{i=1}^{G}\frac{1}{|y_i|}\sum_{t=1}^{|y_i|}
\\
\min\Big(
s_{i,t}(\theta)A_i,\,
\operatorname{clip}\!\big(
s_{i,t}(\theta),
1-\epsilon_{\text{low}}^{\text{seq}},
1+\epsilon_{\text{high}}^{\text{seq}}
\big)A_i
\Big)
\Bigg]
\end{multline}

This sequence-level approach stabilizes training by reducing sensitivity to outlier tokens.
However, all tokens within a sequence share the same ratio and advantage, making it difficult to attribute credit to specific positions.
Additionally, conservative clipping thresholds, required to maintain stability, result in many samples being truncated and slow policy improvement.
Relaxing the thresholds could improve learning but risks reintroducing the instability GSPO seeks to eliminate.

\section{Entropy Importance Sampling Policy Optimization}
ESPO is designed to improve data efficiency and training stability by structuring policy updates according to model uncertainty. At a high level, ESPO consists of two stages.
First, ESPO performs \textit{Entropy Grouping Importance Sampling}, where tokens within each sequence are grouped according to their entropy and the sequence-level importance ratio is decomposed into group-level ratios.
This enables more fine-grained credit assignment by allowing updates to focus on high-uncertainty decision regions while maintaining coherence within low-uncertainty segments.
Second, ESPO applies \textit{Entropy Adaptive Clipping}, which assigns different clipping ranges to different entropy groups, encouraging exploration on uncertain tokens while enforcing stability on confident ones.

We first formulate the complete ESPO objective (Section 3.1), then elaborate on each component: entropy-adaptive clipping (Section 3.2), token grouping by entropy (Section 3.3), and token-level advantage normalization (Section 3.4). 

\subsection{Objective Formulation}

Formally, given a query $x \sim \mathcal{D}$ and a group of $G$ sampled responses 
$\{y_i\}_{i=1}^{G} \sim \pi_{\theta_{\text{old}}}(\cdot|x)$, 
we partition each sequence $y_i$ into a collection of groups $\{\tau\}$, 
where each groups $y_{\tau}$ is not necessarily contiguous in the token space. 
Instead of relying on textual or syntactic boundaries, we group samples based on the model’s internal features that reflect uncertainty. Such uncertainty-related features can be defined at different orders, and ESPO specifically leverage second-order signals(e.g., policy entropy) for grouping.
These features reflect the model’s internal uncertainty and confidence structure, 
allowing the groups to capture regions of similar policy behavior rather than continuous spans in surface form.
\ourpo~defines the following objective:

\begin{multline}
\mathcal{J}_{\mathrm{ESPO}}(\theta) = 
\mathbb{E}_{x \sim \mathcal{D}, \, \{y_i\}_{i=1}^G \sim \pi_{\theta_{\text{old}}}(\cdot|x)} 
\bigg[ 
\frac{1}{G} \sum_{i=1}^G \textcolor{red}{\frac{1}{|\tau|}\sum_{\tau} \frac{1}{|y_\tau|}}\\\sum_{t=1}^{|y_\tau|} 
\min \big( s_{\tau}^{i,t}(\theta)\hat{A}_{\tau}^{i,t}, \,
\operatorname{clip}(s_{\tau}^{i,t}(\theta), 1-\textcolor{red}{\epsilon_\tau}, 1+\textcolor{red}{\epsilon_\tau})\hat{A}_{\tau}^{i,t} 
\big)
\bigg]
\end{multline}
where $s_{\tau}^{i,t}(\theta)$ denotes the importance ratio at the token-level within groups $\tau$, 
$\epsilon_\tau$ is the entropy adaptive clipping threshold, and $\hat{A}_{i,t}$ is the normalized token-level advantage.

\subsection{Entropy Adaptive Clipping}
\label{sec:dynamic_clipping}
Entropy provides a natural and model-intrinsic signal for measure token-level uncertainty.
Following prior studies, we leverage predictive entropy to both token groups and determine their clipping range.

Specifically, we first compute the token-level entropy:
\begin{align}
e_t^i = -\sum_{v \in \mathcal{V}} 
\pi_{\theta_{\text{old}}}(v \mid x, y_{i,<t})
\log \pi_{\theta_{\text{old}}}(v \mid x, y_{i,<t})
\end{align}
where $\mathcal{V}$ denotes the vocabulary space.
The entropy reflects how uncertain the model is about the next-token distribution, with higher values indicating less confidence. 

In \ourpo, we extend this signal to the fine-grained level.
Tokens are grouped into separation $\{\tau\}$ according to their entropy values,
so that tokens with similar uncertainty belong to the same separation:
\begin{align}
\tau_k = \{\, t \mid  e_t^i \in \mathcal{R}_k \,\}, \qquad 
\{\tau_k\}_{k=1}^{K_i} = \mathcal{S}(y_i),
\end{align}
where $\mathcal{R}_k$ denotes the entropy range corresponding to the $k$-th group and $\mathcal{S}$ represents the whole response sentence.
In contrast to previous methods that apply fixed ranges,
\ourpo~uses entropy statistics to dynamically compute each group’s clipping bound:
\begin{align}
\epsilon_\tau = 
 \frac{\alpha}{\log |\mathcal{V}|}\cdot
\frac{ e_t^i}
{|\tau|},
\end{align}

where $\alpha$ is a global scaling factor and the denominator $\log |\mathcal{V}|$ serves as the theoretical upper bound of the token entropy, 
normalizing the clipping coefficient across different vocabulary scales. 
Consequently, groups with higher average normalized entropy obtain larger clipping ranges, 
allowing more freedom for policy updates in uncertain regions, 
while low-entropy, confident groups are constrained by tighter clipping bounds to maintain stability.

This entropy-normalized adaptive design enables \ourpo~to dynamically balance exploration and stability, 
allocating larger optimization flexibility to uncertain regions while preserving GSPO’s stable training behavior across the entire sequence.

\subsection{Entropy Grouping Importance Sampling}
In the same way as GSPO-token decomposes the importance ratio to align optimization with the reward unit, we also decompose the ratio to \textbf{allow token-wise advantage customization}. Specifically, we define the entropy grouping importance ratio as follows:
\begin{align}
s_{\tau}^{i,t}(\theta) :=
sg\!\left[s_{\tau}^i(\theta)\right] 
\cdot 
\frac{\pi_{\theta}(y_{i,t}|x,y_{i,<t})}
     {sg\!\left[\pi_{\theta_{\text{old}}}(y_{i,t}|x,y_{i,<t})\right]},
\end{align}
where $sg[\cdot]$ denotes the stop-gradient operator that blocks backpropagation through the detached component.
The  scaling term at the token grouping level $s_{\tau}^i(\theta)$ is defined as the length-normalized 
sequence ratio:
\begin{align}
s_{\tau}^i(\theta)
=
\left(
\frac{\pi_{\theta}(y_\tau|x)}
     {\pi_{\theta_{\text{old}}}(y_\tau|x)}
\right)^{\!\tfrac{1}{|y_{\tau}|}}
\end{align}

\begin{figure}[h!]
    \centering
    \includegraphics[width=0.5\textwidth]{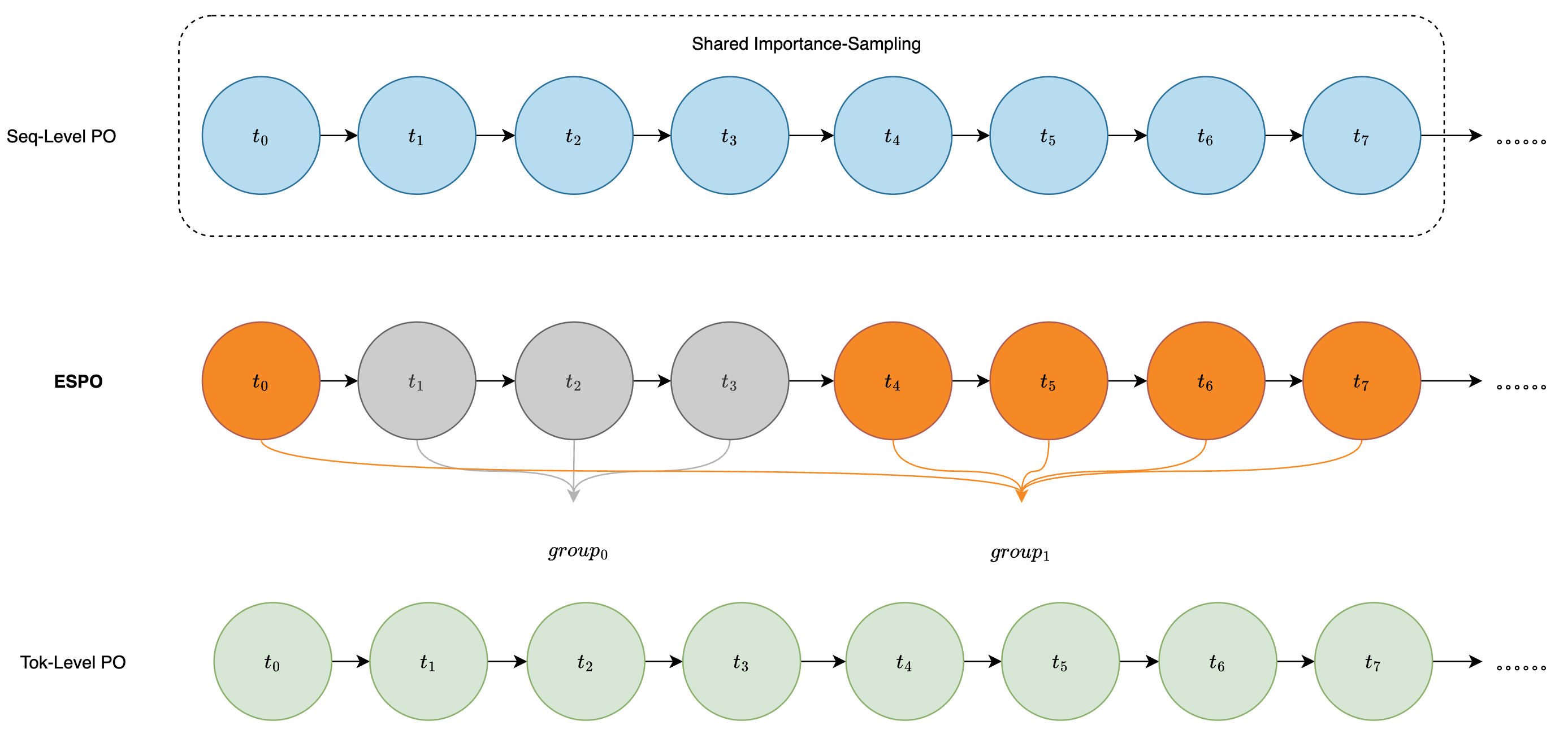}
    \caption{Comparison of Sequence-level, Token-level Algorithms and \ourpo~. \ourpo\ leverages entropy-based decomposition for optimization}
    \label{fig:espo}
\end{figure}

Intuitively, $s_{\tau}^i(\theta)$ captures global consistency throughout the group,
while the token-level ratio $\pi_\theta/\pi_{\theta_{\text{old}}}$ provides fine-grained local sensitivity.
This hierarchical construction allows \ourpo~to adaptively modulate gradients within each group,
preserving stable updates for globally consistent spans and emphasizing local corrections where necessary. As shown in Figure \ref{fig:espo}, \ourpo~introduce entropy-based decomposition, enabling adaptive gradient modulation within each group.
\subsection{Token Grouping Advantage Normalization}
To address the long-standing issue of sparse reward signals and coarse credit assignment in reinforcement learning for large language models, many new works are emerging. For example, \cite{kazemnejad2025vinepporefiningcreditassignment},\cite{zuo2025ttrltesttimereinforcementlearning}, and \cite{liu2025flowgrpotrainingflowmatching} achieve fine-grained credit assignment by using the language model itself to perform a simple value function estimation, resulting in better performance than the baseline. This demonstrates the importance of fine-grained credit assignment in the online reinforcement learning (RL) training of reasoning models.

In this paper, We refine the advantage estimation at a finer granularity and compute token-wise normalized advantages within each group to better reflect local reward contributions.  
Formally, the token-level normalized advantage is defined as

\begin{equation}
\label{eq:token_adv}
\begin{aligned}
\hat{A}_{\tau}^{i,t}
&=
\mathbb{I}[i \in \mathcal{G}]
\cdot
\frac{R(y_i,a) - \mu_{\mathcal{G}}}{\sigma_{\mathcal{G}}}, \\[0.6em]
\text{where}\quad
\mathcal{G}
&:= \{\, i \mid \text{verifier succeeds} \,\}, \\
\mu_{\mathcal{G}}
&:= \operatorname{mean}\!\left(\{R(y_i,a)\}_{i \in \mathcal{G}}\right), \\
\sigma_{\mathcal{G}}
&:= \operatorname{std}\!\left(\{R(y_i,a)\}_{i \in \mathcal{G}}\right).
\end{aligned}
\end{equation}

Here $R(y_i,a)$ denotes the group-aware reward assigned by the verifier.  
Unlike GSPO, where the same advantage is propagated to all tokens in a sequence, \ourpo~allows each token (or group) to be weighted according to its relative contribution to the final reward.  
This design alleviates the sparsity of the reward signal by distributing gradients more precisely to informative regions, 
ensuring that learning is concentrated on tokens that are most responsible for success while suppressing noise from irrelevant or failed samples.  
As a result, \ourpo~enhances both the stability and efficiency of credit assignment across long and complex sequences.

\section{Experiments}

\subsection{Experimental Setup}

\paragraph{Baselines.} 
We compare \ourpo~with two classes of policy optimization baselines: (1) \textbf{sequence-level} methods (GSPO) and (2) \textbf{token-level} methods (DAPO, GMPO, CISPO). This covers the spectrum of granularity choices in contemporary LLM fine-tuning.

\paragraph{Configuration.} 
All experiments use the VERL framework~\cite{sheng2024hybridflow} with vLLM~\cite{kwon2023efficient} for rollouts. We train on SimpleRL~\cite{zeng2025simplerl} (8,192 prompts) for 3 epochs (200 steps) with a fixed learning rate of $1\times10^{-6}$. Each prompt generates $G=8$ responses (temperature 1.0, max length 16,384) with batch size 128. Training runs on a 32-GPU H100 cluster.

\paragraph{Evaluation.} 
We report accuracy on MATH500~\cite{lightman2023lets} and average@32 on AIME 2024/2025~\cite{aime24,aime25} and HMMT~\cite{balunovic_srimatharena_2025}. All methods are evaluated on Qwen3 base models of varying scales and architectures: dense (1.7B, 4B, 14B) and MoE (30B-A3B), under identical initialization and hyperparameters.

\begin{table*}[t]
\centering
\small
\setlength{\tabcolsep}{10pt}
\begin{tabular}{lccccc}
\toprule
\multicolumn{1}{c}{\textbf{Algorithm}} &
\makecell[c]{\textbf{AIME24} \\ (Avg@32)} &
\makecell[c]{\textbf{AIME25} \\ (Avg@32)} &
\makecell[c]{\textbf{MATH500} \\ (Avg@1)} &
\makecell[c]{\textbf{HMMT} \\ (Avg@32)} &
\textbf{AVG} \\
\midrule

\rowcolor{groupgray}
\multicolumn{6}{c}{\textbf{\textit{Qwen3-1.7B-Base}}} \\
DAPO                     & 11.9 & 7.8 & 70.0 & 1.5 & 22.8 \\
GSPO                     & 8.5  & 5.0 & 69.2 & 1.0 & 20.9 \\
GMPO                     & 6.6  & 4.8 & 69.2 & 1.0 & 20.5 \\
CISPO            & \textbf{12.6} & 6.0 & \textbf{71.0} & \textbf{2.4} & 23.0 \\
\rowcolor{ourblue}
\ourpo (Ours)             & \textbf{12.6} & \textbf{9.1} & \textbf{71.0} & 1.8 & \textbf{23.6} \\

\midrule
\rowcolor{groupgray}
\multicolumn{6}{c}{\textbf{\textit{Qwen3-4B-Base}}} \\
DAPO                     & 21.8 & \textbf{22.8} & 85.0 & 10.1 & 34.9 \\
GSPO                     & 19.1 & 18.8 & 85.0 & 8.3 & 33.0 \\
GMPO                     & 20.9 & 20.0 & 83.8 & 8.8 & 33.4 \\
CISPO            & \textbf{24.8} & 20.0 & 85.0 & \textbf{10.8} & 35.1 \\
\rowcolor{ourblue}
\ourpo (Ours)             & 23.3 & 21.4 & \textbf{85.8} & 10.4 & \textbf{35.2} \\

\midrule
\rowcolor{groupgray}
\multicolumn{6}{c}{\textbf{\textit{Qwen3-14B-Base}}} \\
DAPO                     & 35.1 & 25.5 & 89.0 & 14.8 & 41.1 \\
GSPO                     & 33.2 & 28.6 & 89.0 & 14.8 & 41.4 \\
GMPO                     & 33.2 & 25.5 & 90.2 & 13.1 & 40.5 \\
CISPO            & \textbf{41.8} & 30.9 & 90.2 & \textbf{16.0} & 44.7 \\
\rowcolor{ourblue}
\ourpo (Ours)             &39.3 & \textbf{32.9} & \textbf{91.0} & \textbf{16.0} & \textbf{44.8} \\

\midrule
\rowcolor{groupgray}
\multicolumn{6}{c}{\textbf{\textit{Qwen3-30B-A3B-Base}}} \\
DAPO                     & 27.6  & 15.6 & 85.8 & 4.4  & 33.2 \\
GSPO                     & 28.1 & 15.6 & 85.8 & 4.4  & 33.4 \\
GMPO                     & 26.0 & 13.6 & 86.2 & 4.9 & 32.7 \\
CISPO            & 37.1 & 23.2 & 89.0 & \textbf{15.3} & 41.1 \\
\rowcolor{ourblue}
\ourpo (Ours)             & \textbf{40.2} & \textbf{24.6} & \textbf{90.2} & 14.9 & \textbf{42.5} \\
\bottomrule
\end{tabular}
\caption{Main results on mathematical reasoning benchmarks. We report the Avg@32 accuracy for AIME and HMMT datasets, and accuracy for MATH500. The AVG column denotes the macro-average performance across all evaluated tasks, highlighting the comprehensive superiority of \ourpo.}
\label{main_results}
\end{table*}

\begin{figure*}[t]
    \centering

    \begin{subfigure}[t]{0.32\linewidth}
        \centering
        \caption{Training Accuracy}
        \label{ta}
        \includegraphics[width=\linewidth]{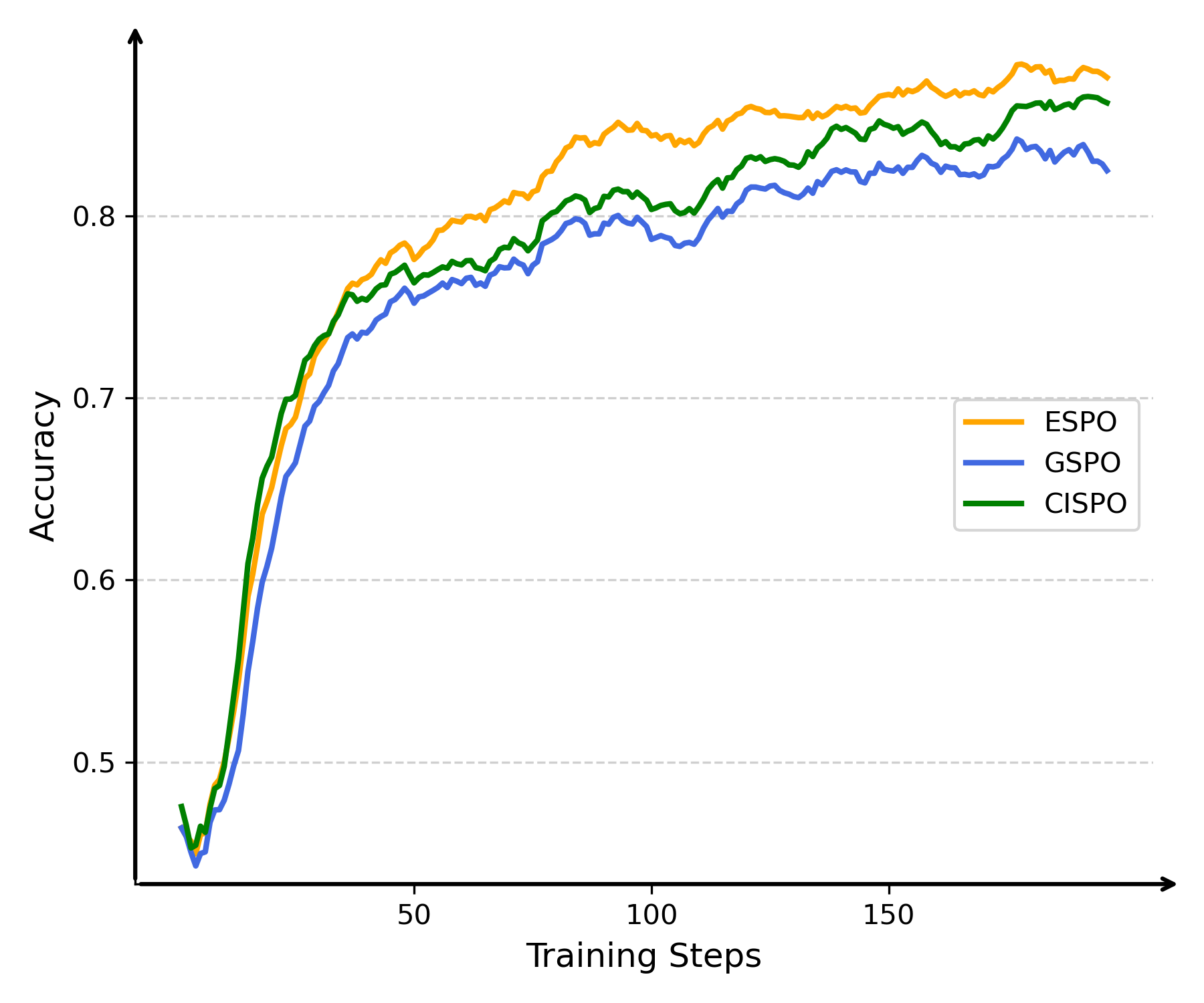}
    \end{subfigure}
    \hfill
    \begin{subfigure}[t]{0.32\linewidth}
        \centering
        \caption{Mean Reponse Length}
        \label{rl}
        \includegraphics[width=\linewidth]{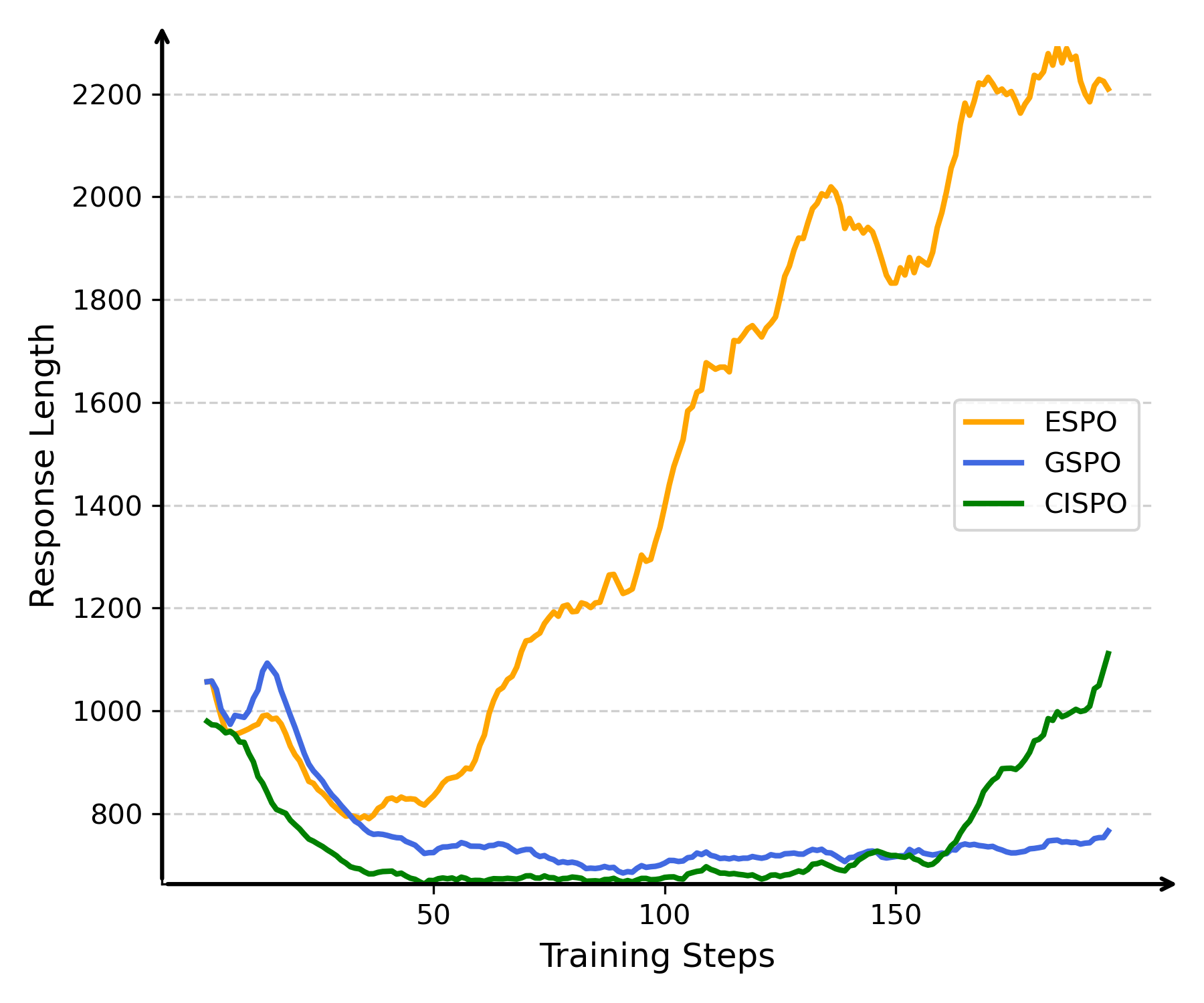}
    \end{subfigure}
    \hfill
    \begin{subfigure}[t]{0.32\linewidth}
        \centering
        \caption{Entropy}
        \label{ent}
        \includegraphics[width=\linewidth]{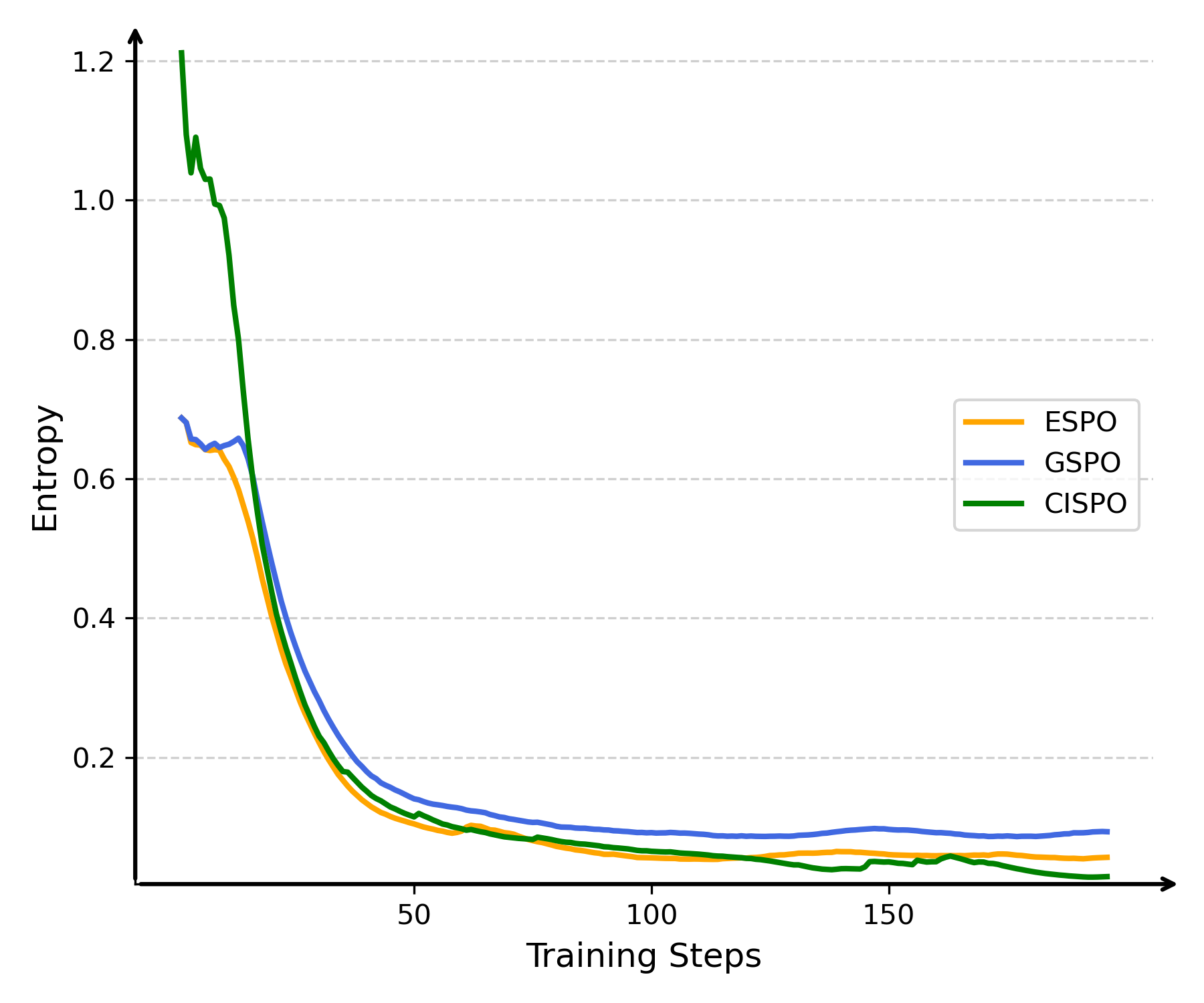}
    \end{subfigure}

    \caption{
    (a) Average accuracy across benchmarks over training steps.
(b) Mean response length as a function of training steps.
(c) Policy entropy variation during training.
All curves are reported under the same training configuration on Qwen3-30B-A3-Base
.}
    \label{fig:1x3}
\end{figure*}

\subsection{Overall Performance Comparison}

\paragraph{Comprehensive Benchmark Results.}
Table~\ref{main_results} presents the performance of ESPO and baseline methods across four mathematical reasoning benchmarks and multiple model scales. 
ESPO achieves the highest average performance in all settings, demonstrating its consistent superiority.

On the smallest dense model (Qwen3-1.7B), ESPO matches or exceeds the best token-level baselines on AIME24 and MATH500 while providing a noticeable 3.1-point improvement on AIME25. 
This indicates that ESPO's fine-grained yet stable optimization is particularly beneficial when model capacity is limited.

With medium-scale dense models (Qwen3-4B and 14B), ESPO maintains a narrow but consistent lead over strong token-level competitors like CISPO. 
While the absolute margins decrease as models become more capable, ESPO still attains the best average scores, suggesting it offers a robust trade-off between granularity and stability. 

The advantage becomes most pronounced on the large MoE model (Qwen3-30B-A3B). 
Here, ESPO outperforms both sequence-level and token-level baselines by substantial margins—achieving up to 12.1-point gains over GSPO on AIME24 and 1.4-point gains over CISPO on AIME25. 
This highlights ESPO's effectiveness in heterogeneous architectures where fixed-granularity methods struggle to adapt.

Across all model sizes and architectures, ESPO delivers balanced improvements without task-specific tuning, confirming its value as a general-purpose optimization strategy.

\paragraph{Training Dynamics.}
Figure~\ref{fig:1x3} compares the training behavior of ESPO with representative sequence-level (GSPO) and token-level (CISPO) methods on Qwen3-30B-A3B. 
Three key observations emerge:

First, ESPO converges at a comparable or faster rate than baselines in terms of training accuracy (Fig.~\ref{fig:1x3}a), indicating more efficient optimization. 
Second, while baseline methods stabilize or only slowly increase their response lengths, ESPO sustains a steady growth in generated sequence length throughout training (Fig.~\ref{fig:1x3}b), demonstrating continued exploration of longer reasoning trajectories. 
Third, ESPO demonstrates more stable entropy reduction than CISPO while starting from a more moderate initial value than both CISPO  and GSPO (Fig.~\ref{fig:1x3}c). Its smooth descent leads to a final policy that is marginally more confident than GSPO's, indicating effective exploration without excessive randomness.

Collectively, these dynamics suggest that ESPO optimizes more efficiently (faster accuracy rise), explores more thoroughly (longer trajectories), and refines its policy more stably (smooth entropy decay) than fixed-granularity baselines.

\subsection{Analysis of ESPO's Mechanisms}
\paragraph{Addressing Gradient Underutilization.}
\begin{figure}[h!]
    \centering
    \includegraphics[width=0.95\linewidth]{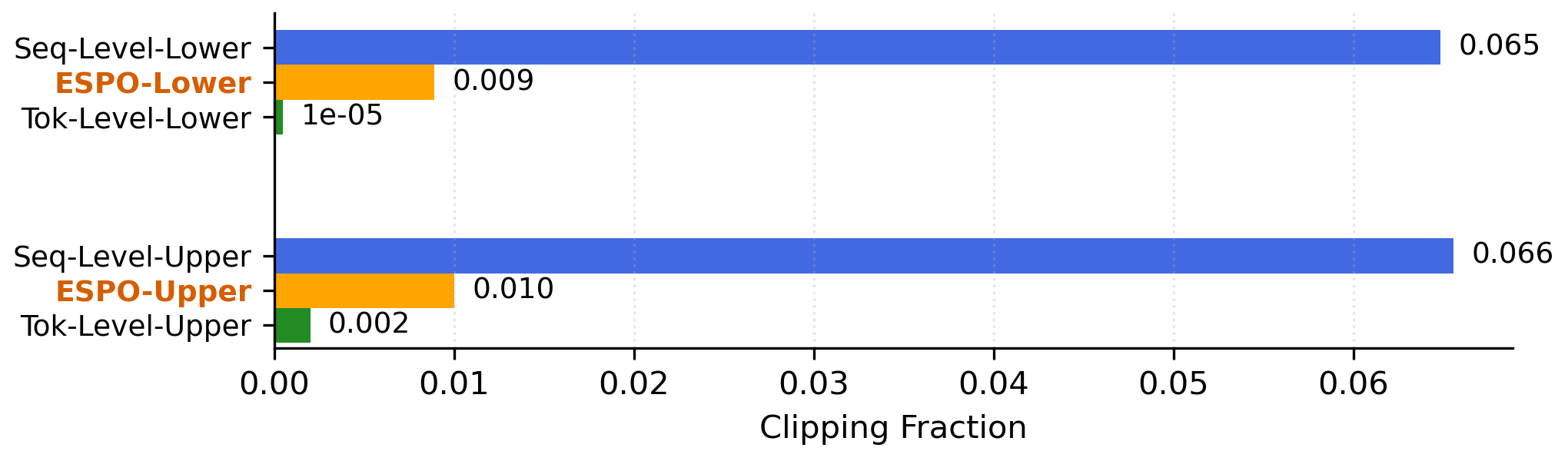}
    \caption{Comparison of Sequence-level, Token-level Algorithms and \ourpo~Clip Fraction}
    \label{fig:clip_fraction_compact_v2}
\end{figure}
The performance advantages of ESPO stem from its ability to overcome the core limitations of fixed-granularity optimization. Figure~\ref{fig:clip_fraction_compact_v2} compares the clipping fractions for both upper and lower bounds across methods, providing insights into their update efficiency and stability.

\textbf{Sequence-level methods} exhibit the highest clipping fractions, indicating that a large portion of samples are truncated during updates. This occurs because a single sequence-level importance ratio is shared across all tokens; once this ratio exceeds the clipping threshold, the entire trajectory is discarded. Consequently, many informative gradients are suppressed, leading to the \textit{gradient underutilization} problem highlighted in our introduction.

\textbf{Token-level methods} alleviate excessive truncation by using wider clipping ranges and finer-grained importance ratios. However, this relaxed constraint permits updates with highly variable token-wise ratios to pass through, introducing substantial noise into gradient estimates. Such instability is particularly problematic when importance ratios vary sharply across tokens, undermining training robustness.

\textbf{ESPO} maintains low clipping fractions without resorting to overly permissive ranges. By grouping tokens based on entropy and applying group-aware clipping, ESPO selectively constrains high-variance updates while preserving informative gradients. This approach enables more efficient use of training samples, avoiding both the wastefulness of sequence-level methods and the instability of token-level alternatives.

\paragraph{Validating Adaptive Optimization.}
\begin{figure}[h!]
    \centering
    \includegraphics[width=0.85\linewidth]{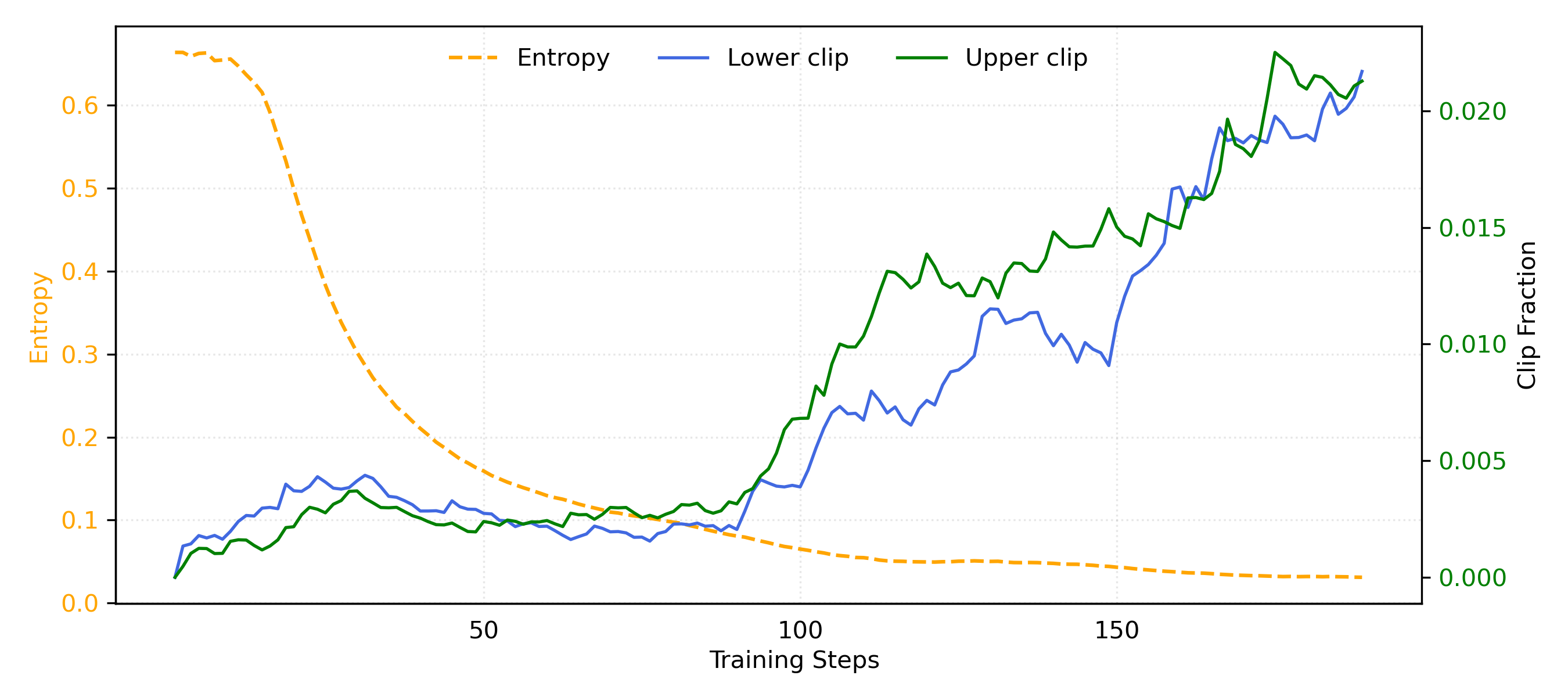}
    \caption{Comparison of Entropy and Clip Fraction}
    \label{fig:entropy}
\end{figure}
ESPO achieves this balanced behavior through its entropy-adaptive clipping mechanism. Figure~\ref{fig:entropy} validates that this mechanism operates as designed, dynamically aligning optimization strength with model uncertainty.

During early training stages, the policy exhibits high predictive entropy (orange dashed line), reflecting broad uncertainty about token predictions. In this phase, ESPO assigns wider clipping ranges (lower clipping fractions), allowing greater exploration and policy deviation. As learning progresses, entropy steadily decreases as the model gains confidence, and ESPO automatically tightens its clipping constraints—indicated by the rising clipping fractions (blue and green lines). This adaptive behavior ensures that more samples fall within the trusted region once the policy stabilizes.

The inverse relationship between entropy and clipping fraction confirms that ESPO successfully modulates exploration intensity according to the model's own uncertainty: high entropy encourages learning diversity, while low entropy enforces conservative updates. This principled adaptation explains how ESPO maintains stable optimization throughout training while still exploring effectively.

\begin{table*}[t]
\centering
\small
\setlength{\tabcolsep}{10pt}
\begin{tabular}{lccccc}
\toprule
\multicolumn{1}{c}{\textbf{Algorithm}} &
\makecell[c]{\textbf{AIME24} \\ (Avg@32)} &
\makecell[c]{\textbf{AIME25} \\ (Avg@32)} &
\makecell[c]{\textbf{MATH500} \\ (Avg@1)} &
\makecell[c]{\textbf{HMMT} \\ (Avg@32)} &
\textbf{AVG} \\
\midrule

\textbf{\ourpo} 
& \textbf{40.2} 
& \textbf{24.6} 
& \textbf{90.2} 
& \textbf{14.9} 
& \textbf{42.5} \\

\midrule
-- w/o Adaptive Clipping-1 (0.0003,0.0004)
& 32.4 & 15.3 & 87.2 & 6.1 & 35.3 \\

-- w/o Adaptive Clipping-2 (0.2,0.28)
& 27.1 & 17.9 & 87.0 & 8.2 & 35.1 \\

-- w/o Token Grouping
& 25.9 & 15.2 & 84.8 & 3.7 & 32.4 \\
\bottomrule
\end{tabular}
\caption{Ablation results of ESPO on AIME24, AIME25, MATH500, and HMMT.
Each row corresponds to removing a specific component from ESPO, demonstrating its contribution to overall performance across different benchmarks.}
\label{tab:acc8_results}
\end{table*}
\begin{table*}[t]
\centering
\small
\setlength{\tabcolsep}{10pt}
\begin{tabular}{lccccc}
\toprule
\multicolumn{1}{c}{\textbf{Alpha}} &
\makecell[c]{\textbf{AIME24} \\ (Avg@32)} &
\makecell[c]{\textbf{AIME25} \\ (Avg@32)} &
\makecell[c]{\textbf{MATH500} \\ (Avg@1)} &
\makecell[c]{\textbf{HMMT} \\ (Avg@32)} &
\textbf{AVG} \\
\midrule

$\alpha$ = 0.01
& 37.3 & 23.2 & 86.4 & 14.8 & 40.4 \\

$\alpha$ = 0.02
& \textbf{40.2} & \textbf{24.6} & \textbf{90.2} & \textbf{14.9} & \textbf{42.5} \\

$\alpha$ = 0.05
& 36.6 & 23.2 & \textbf{90.2} & 14.1 & 41.0 \\

$\alpha$ = 0.10
& 38.2 & 24.4 & 89.8 & 13.6 & 41.5 \\
\bottomrule
\end{tabular}
\caption{Sensitivity analysis of ESPO with respect to the hyperparameter $\alpha$ on AIME24, AIME25, MATH500, and HMMT benchmarks.
We vary $\alpha$ while keeping all other components fixed and report the resulting performance across benchmarks.}
\label{tab:alpha_results}
\end{table*}

\subsection{Ablation Studies}
Having established that ESPO's adaptive mechanism effectively addresses the limitations of fixed-granularity methods, we now validate the necessity of its individual design choices through ablation experiments. All ablations use the Qwen3-30B-A3B model under identical training configurations.

\paragraph{Core Components.}
We first examine the contribution of ESPO's two key innovations: adaptive clipping and entropy grouping importance sampling. Table~\ref{tab:acc8_results} shows that removing either component significantly degrades performance.

\textbf{Removing adaptive clipping}—by replacing it with fixed clipping schemes—consistently harms results. We test two variants: (1) aggressive sequence-level clipping (0.0003, 0.0004) and (2) conservative token-level clipping (0.2, 0.28). Both cause comparable performance drops despite their different granularities, confirming that static clipping cannot adapt to evolving training dynamics and that ESPO's adaptive mechanism is essential.

\textbf{Removing entropy grouping} causes the most severe degradation across all benchmarks. Without structured grouping based on entropy, clipping operates in an uncoordinated token-wise manner, significantly increasing gradient variance. This demonstrates that merely adjusting clipping ranges is insufficient; aligning clipping behavior with the model's entropy structure is crucial for stable optimization.

The substantial performance gap between the full ESPO and these ablated versions validates that both components are necessary and work synergistically.

\paragraph{Grouping Proportion.}
\begin{figure}[h!]
    \centering
    \includegraphics[width=0.8\linewidth]{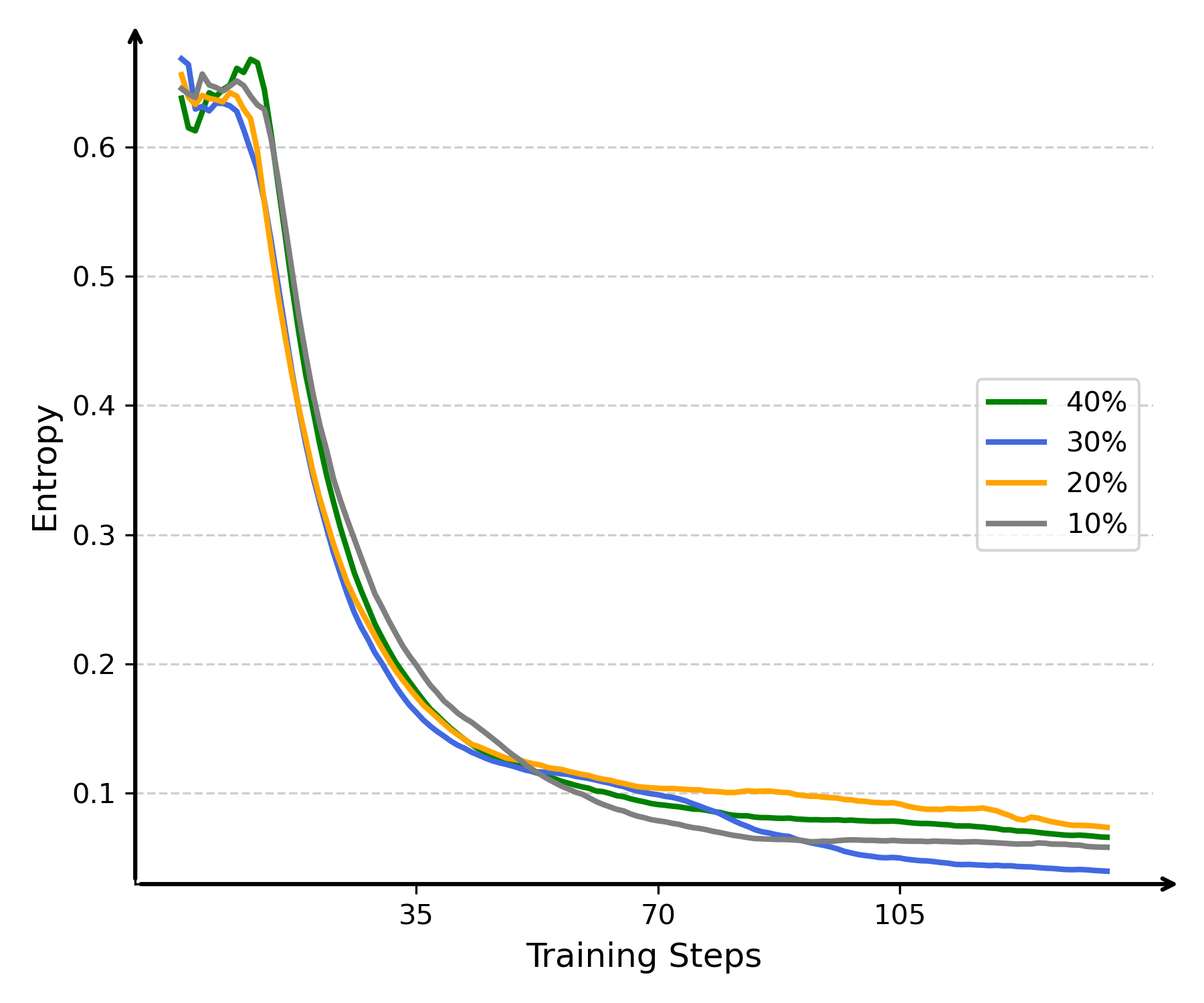}
    \caption{Comparison of different high entropy percentage}
    \label{fig:rho_ent}
\end{figure}
Given the importance of token grouping, we analyze the optimal proportion of high-entropy tokens to include. Figure~\ref{fig:rho_ent} reveals a clear trade-off.

Selecting too few tokens (e.g., 10\%) leads to weaker exploration. As observed in prior work~\cite{wang2025beyond}, overly aggressive filtering can remove certain useful tokens that still correspond to meaningful decision points, resulting in faster entropy decay and reduced diversity early in training. Consequently, the model may converge prematurely before sufficiently exploring the reasoning space.

Conversely, selecting too many tokens (e.g., 40\%) introduces a substantial number of low-entropy tokens. These tokens are often highly certain and knowledge-driven, and contribute limited information for improving reasoning behavior. Including them in large proportions dilutes the influence of informative gradients and reduces exploration efficiency.

Across experiments, selecting approximately 20\% of the highest-entropy tokens provides the optimal balance. This intermediate choice preserves enough informative decision points to sustain exploration while avoiding excessive inclusion of low-entropy tokens that can hinder optimization, validating our grouping strategy.

\paragraph{Hyperparameter Sensitivity.}
Finally, we examine the sensitivity of ESPO to $\alpha$, the global scaling factor controlling update strength (Table~\ref{tab:alpha_results}). Performance peaks at $\alpha = 0.02$, with both smaller and larger values yielding suboptimal results.

When $\alpha$ is too small (0.01), clipping becomes overly aggressive, limiting exploration. When $\alpha$ is too large (0.05, 0.10), clipping becomes too permissive, reducing stability. This narrow optimal range underscores the importance of balanced exploration and confirms that our chosen $\alpha = 0.02$ represents an effective default setting.


\section{Conclusion}

We introduce Entropy Importance Sampling Policy Optimization (ESPO), an entropy-aware RL framework that groups tokens by uncertainty and clips adaptively. On challenging mathematical benchmarks, ESPO exhibits accelerated convergence, reduced clipping, and superior sample-efficiency versus strong baselines under equal compute. Future directions include alternative grouping criteria, learnable grouping mechanisms, and extensions to multi-turn dialogue and program synthesis.

\appendix

\section*{Ethical Statement}

All experiments were conducted in accordance with the applicable institutional and national guidelines on research involving (open-source/public) data.
The dataset(s) used in this study are publicly available and were originally released under permissive licenses that allow academic reuse.
Our models follow standard practices for large-scale language model training; they may still produce biased, offensive, or factually incorrect outputs.

\bibliographystyle{named}
\bibliography{espo}

\end{document}